\documentclass{article}

\PassOptionsToPackage{numbers}{natbib}
\usepackage[position,final]{neurips_2026}

\usepackage[utf8]{inputenc} % allow utf-8 input
\usepackage[T1]{fontenc}    % use 8-bit T1 fonts
\usepackage{url}            % simple URL typesetting
\usepackage{booktabs}       % professional-quality tables
\usepackage{amsfonts}       % blackboard math symbols
\usepackage{nicefrac}       % compact symbols for 1/2, etc.
\usepackage{microtype}      % microtypography
\usepackage{xcolor}         % colors
\usepackage{bm}
\usepackage{graphicx}
\usepackage{soul}
\newcommand{\supp}{\operatorname{supp}}
\usepackage{amsmath,amssymb,amsthm,mathtools}
\usepackage{algorithm}
\usepackage{algpseudocode}
\usepackage{tabularx}
\usepackage{ragged2e}
\usepackage{array}
\usepackage{multirow}   % \multirow
\usepackage{graphicx}   % \rotatebox
\usepackage{enumitem}
\usepackage{array}
\usepackage{xcolor}
\definecolor{darkblue}{HTML}{627EBC}
\definecolor{darkred}{HTML}{C0483A}
\usepackage[breaklinks,colorlinks=true,citecolor=darkblue,linkcolor=darkred,urlcolor=darkblue]{hyperref}

\newcommand{\Pmask}{\mathbf{P}_{\bm{\alpha}}}
\newcommand{\simplex}{\Delta}
\DeclareMathOperator*{\argmin}{arg\,min}
\DeclareMathOperator*{\vertset}{vertices}

\usepackage[most]{tcolorbox}
\tcbuselibrary{skins,breakable}
\usepackage{enumitem}

\definecolor{plain}{HTML}{eaaf47}
\definecolor{persona}{HTML}{627ebc}
\definecolor{ToM}{HTML}{9b5d92}

\newcommand{\cplain}{\textbf{\textcolor{plain}{Plain}}}
\newcommand{\cpersona}{\textbf{\textcolor{persona}{Persona}}}
\newcommand{\ctom}{\textbf{\textcolor{ToM}{ToM}}}

\title{Claims of emergence in AI should be grounded in information decomposition}
\author{%
  Christoph Riedl\\%\thanks{Use footnote for providing further information about author (webpage, alternative address)} \\
  Khoury College of Computer Sciences\\
  Northeastern University\\
  Boston, MA 02115 \\
  \texttt{c.riedl@northeastern.edu} \\
  \And
  Fernando E. Rosas\\
  Department of Informatics\\
  University of Sussex\\
  Brighton, BN1 9RH\\
  \texttt{f.rosas@sussex.ac.uk}
}

\begin{document}

\maketitle

\begin{abstract}
When a language model develops capabilities not traceable to its individual components, when a multi-agent AI system outperforms each agent on its own, or when a human-AI pair succeeds where either fails in isolation, AI researchers describe it as a case of ``synergy'' or ``emergence''. However, these are usually informal assessments not supported by a theoretical framework, making these claims impossible to compare or even meaningfully dispute. Here we argue that \emph{information decomposition can serve as a unifying formal framework to operationalize assessments about synergy in AI systems}. Information decomposition provides a way to distinguish qualitatively different \emph{types} of information, including unique information available from individual components, redundant information shared across multiple components, and synergistic information accessible only from parts taken together. This gives claims of emergence a precise, falsifiable criterion: a system's behavior is emergent only if it carries synergistic information about task-relevant outcomes. We apply this lens to attention heads, autonomous agents, and human-AI teams, showing how it organizes existing knowledge across scales and substrates, while also generating cross-level conjectures that siloed fields would miss.
\end{abstract}

\section{Introduction}

AI research describes some of its most interesting phenomena in the language of \emph{emergence}. Transformer models develop internal representations irreducible to single neurons \citep{elhage2022toy,urbina2026brain,wang2023interpretability}; multi-agent systems solve tasks that no single agent can \citep{nielsen2026learning,riedl2026emergent,qian2024chatdev}; and human-AI teams succeed where either partner fails alone \citep{riedl2026quantifying,mozannar2020consistent,bansal2021does}. These cases differ in scale and substrate, yet make the same structural claim: the capabilities of the whole became ``more than the sum of its parts.'' 

Such ``more-than-the-sum'' effects are often described as \emph{synergy}. Yet, in current AI research, ``synergy'' is an umbrella term encompassing several different ideas. 
In mechanistic interpretability, it may refer to interaction effects between attention heads, non-localized circuits across layers, or superposition of representations \citep[e.g.,][]{nanda2023emergent,grant2025emergent,de2020decisions}. In multi-agent systems, synergy corresponds to ensemble gain, enabled by coordination, over what independent agents can achieve \citep[e.g.,][]{chen2023agentverse,nielsen2026learning,qian2024chatdev}. 
In human-AI systems, synergy is tied to substitution, elicitation, prompting, and learning effects \citep[e.g.,][]{vaccaro2024combinations,yuan2026towards,meng2026exploring}. 
These uses are related but not equivalent, and the differences are not mere semantics but a barrier to scientific progress: without a shared formal criterion for what synergy means, these claims cannot be compared, aggregated, or even meaningfully disputed.

In this paper, we argue that Partial Information Decomposition (PID)~\cite{williams2010nonnegative} provides conceptual and computational tools that can organize these disparate notions of synergy into a common, unifying framework. 
PID proposes to decompose the information that multiple sources carry about a target into information unique to one source, information redundantly shared across sources, and synergistic information available only from the sources jointly (Figure \ref{fig:decomposition}a). This gives emergence a precise, mereological meaning: a system's collective behavior is emergent only if it carries synergistic information \citep{rosas2020reconciling}. Unlike the scale-dependent ``emergent abilities'' of \citet{wei2022emergent}, which can be artifacts of the evaluation metric \citep{schaeffer2023emergent}, this notion applies to any system of interacting components and does not hinge on how performance is scored.
Crucially, by treating unique, redundant, and synergistic information as \emph{fundamental types of information}, PID offers a substrate-agnostic calculus revealing common constraints that drive how information is integrated, whether across silicon neurons, autonomous agents, or human-AI teams. This, in turn, suggests a general pressure for increasingly capable systems to transition from robust redundancy to efficient synergy under resource constraints.

\textit{We propose PID---applied either as an exact computation or as a conceptual scaffold---as a framework to conceptualize collective intelligence across scales (within and across agents) and substrates (artificial and human).}%
\footnote{On collective intelligence in human groups, see \citet{riedl2021quantifying}; for a related multiscale view in biology, see \citet{mcmillen2024collective}.}
We argue that PID explains---rather than replaces---existing metrics. 
Different communities already compute synergy-like quantities---ensemble gains \citep{qian2024chatdev}, local (point-wise) surprisal \citep{ince2017measuring,westby2023collective}, residuals over independent-source null models \citep{lee2025complex}, mechanistic null models \citep{riedl2026gtbm}, 
latent variable models \citep{riedl2026quantifying}, 
variance-component decompositions of dyadic data~\citep{kenny1984social}, and 
compression-based measures~\citep{cilibrasi2005clustering}---and PID interprets each as a \textit{partial projection} onto the same underlying decomposition, clarifying what each identifies even when full PID estimation is intractable.

% #################################
% ## Theory
% #################################
\section{Partial Information Decomposition as a General Lens on Interdependence}\label{sec:PID}

Here we provide a brief introduction to the mathematical formalism of Partial Information Decomposition (PID) \citep{williams2010nonnegative} and some of its recent extensions \citep{mediano2025toward}. While this section introduces PID as a mathematical instrument, we also highlight where it serves as a conceptual framework.

\subsection{Synergy}\label{sec:primer_synergy}
\begin{figure}
    \centering
    \includegraphics[width=.85\textwidth]{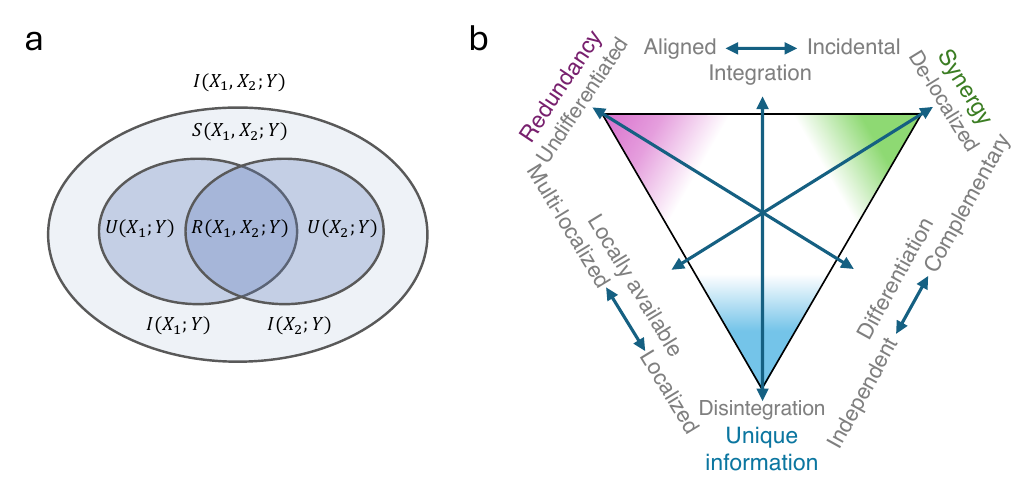}
    \vspace{-8pt} 
    \caption{
        \textbf{a)} Information decomposition with two source variables $X_1$ and $X_2$, and target variable $Y$.
        \textbf{b)} The PID conceptual triangle resolves ambiguities in system dynamics. Traditional dichotomies in complex systems literature often conflate mathematically distinct properties. Integration conflates redundancy (integration-as-oneness) with synergy (complementary cooperation). Differentiation conflates unique information (strict independence) with synergy (interactive differentiation). Furthermore, locality conflates unique (singularly local) with redundant (multiply local) information, whereas synergistic representations are strictly de-localized (adapted from \citet{luppi2024information}).
        }
    \label{fig:decomposition}
    \label{fig:triangle}
\end{figure}

PID focuses on a deceptively simple question: given a collection of variables representing the parts of a system% (e.g., neurons in a MLP, or agents in a multi-agent system)
, how is the information they contain %have about a particular outcome 
distributed among them? Concretely,%The below provides an exhaustive partition of every bit of information the system carries: 
\begin{itemize}[leftmargin=2em]
    \item how much of that information is contained in individual variables alone (unique information), 
    \item how much is contained in several different variables (redundancy), and 
    \item how much is contained in groups of variables without being accessible from any subset of them (synergy)? 
\end{itemize}
Formally, PID considers source variables $X_1, \ldots, X_n$ that hold information about a target variable $Y$ (Figure~\ref{fig:decomposition}a). For example, $X_i$ could represent behavioral time-series of agents in a multi-agent system and $Y$ their performance. The total information that the sources have about the target is captured by Shannon’s mutual information $I(X_1, \ldots, X_n; Y)$ %, which measures the amount of information that all sources have about the target 
\citep{shannon1948mathematical}. 
% CR: trimming some stuff that seems too basic for NeurIPS audience
% Mutual information is a non-linear generalization of correlation, which for Gaussian variables reduces to the log-ratio of total variance to residual variance:
%
% \begin{equation}
%    I(X_1, \dots, X_n;\, Y) = \tfrac{1}{2}\log\lvert\operatorname{Var}(Y)\rvert - \tfrac{1}{2}\log\lvert\operatorname{Var}(Y \mid X_1, \dots, X_n)\rvert,
%    \label{eq:mi_gaussian}
%\end{equation}
%
% being closely related to the explained variance in multiple linear regression.
For $n{=}2$ sources, the joint mutual information decomposes via the chain rule of information into $I(X_1,X_2;Y) = I(X_1;Y)+I(X_2;Y|X_1)$, where the first term captures the information about $Y$ given by $X_1$, and the second the additional information given by $X_2$~\cite{cover1991elements}. 
PID proposes to further decompose these two components into four \emph{information atoms}~\cite{williams2010nonnegative}:
\begin{equation}
    I(X_1, X_2;\, Y) = \underbrace{%
  \mathrm{Red}(X_1, X_2;\, Y) + \mathrm{Un}(X_1;\, Y | X_2)%
}_{I(X_1;Y)}
 + \underbrace{
 \mathrm{Un}(X_2;\, Y | X_1) + \mathrm{Syn}(X_1, X_2;\, Y)}_{I(X_2;Y|X_1)}.
    \label{eq:pid}
\end{equation}
%
%Redundancy $\mathrm{Red}(X_1, X_2;\, Y)$ captures the information that both sources share about the target. 
This gives three equations and four unknowns. Establishing how to calculate one of the above atoms determines the others. A common choice is to specify their redundancy, $\mathrm{Red}(X_1, X_2;\, Y)$, via the minimum mutual information (MMI) criteria \citep{barrett2015exploration}:
\begin{equation}
    \mathrm{Red}(X_1, X_2;\, Y) =\min\{I(X_1; Y),\, I(X_2; Y)\}.
    \label{eq:MMI}
\end{equation}
While this operationalization is canonical for Gaussian variables~\cite{barrett2015exploration}, it may not be appropriate for more general settings---but can still be used as an upper bound for redundancy~\cite{rosas2016understanding} (see Appendix \ref{sup:technical_details} for details). The PID literature has proposed several alternatives for addressing more general settings, driven by inspiration from different fields such as neuroscience, cryptography, or game theory, each with their own specific desiderata and motivations~\citep{rosas2020operational,kolchinsky2022novel,finn2018pointwise,ince2017measuring,makkeh2018broja,makkeh2021introducing}. 
While many of these formulations have been shown to agree in practical scenarios~\cite{rosas2020operational}, some of their driving desiderata are mutually incompatible~\cite{rauh2014reconsidering}. This suggests that none of them is universally preferable~\cite{james2019unique}, and hence practitioners need to choose between operationalizations based on domain knowledge, and properties of the data and problem of interest (e.g., some formulations are only defined for discrete data). For a detailed comparison of PID operationalizations, see~\cite{liardi2026mathematical}.

% Each source's unique information is then the non-redundant portion of its marginal contribution:
%
% \begin{equation}
%     \mathrm{Un}(X_1;\, Y \setminus X_2) = I(X_1;\, Y) - \mathrm{Red}(X_1, X_2;\, Y),
%     \label{eq:unique}
% \end{equation}
%
% and likewise for $X_2$. Finally, synergy is the information available only when both sources are observed jointly, obtained as the residual:
%
% \begin{equation}
%    \mathrm{Syn}(X_1, X_2;\, Y) = I(X_1, X_2;\, Y) - \bigl[\mathrm{Red}(X_1, X_2;\, Y) + \mathrm{Un}(X_1;\, Y \setminus X_2) + \mathrm{Un}(X_2;\, Y \setminus X_1)\bigr].
%    \label{eq:synergy}
%\end{equation}

For $n\geq 3$ source variables the decomposition is much richer, including redundancies and synergies terms of varying orders~\citep{williams2010nonnegative}. 
In fact, the number of information atoms grows super-exponentially with the number of sources, making the full decomposition intractable for six or more variables~\cite{jansma2025fast}. Four practical strategies have been developed to circumvent this issue and many are accessible through software resources (see Table \ref{tab:software}): 
\begin{itemize}
\item[(i)]~targeting specific atoms of interest rather than calculating the full decomposition~\cite{jansma2025fast}; 
\item[(ii)]~evaluating reduced decompositions that scale linearly in the number of sources
\citep{rassouli2019data,rosas2018information};
\item[(iii)]~computing the full decomposition over small subsets (e.g., all triplets) and aggregating the resulting atoms across subsets \citep{riedl2026emergent,ehrlich2023a}; or 
\item[(iv)]~computing scalar summaries, such as the O-information 
\citep{faes2022new,rosas2019quantifying,stramaglia2021quantifying}, which quantify the net balance between synergy and redundancy without explicitly disentangling them~\cite{rosas2025characterising} (biases in O-information estimation are discussed in \citep{gehlen2024bias}).
\end{itemize}

% ######################
% ## TRIANGLE FIGURE
% ######################
Beyond formal definitions and computational techniques, the distribution of information across PID atoms dictates a fundamental architectural tradeoff that can be used to study a system's representational capacity and structural robustness~\cite{luppi2024information,gutknecht2025shannon,riedl2026emergent}: \emph{redundant information provides robustness (since it is found in multiple signals) while synergistic information provides efficiency}. As the number of sources increases, the synergistic portion of the decomposition expands combinatorially---the count of higher-order synergy atoms grows super-exponentially, giving synergy far greater representational capacity than redundant or unique information for a given number of components \citep{rosas2020operational,gutknecht2025shannon}. However, since it depends on joint contributions from multiple sources, synergy is fragile and can be disrupted by the failure of even a single element (however, see also \cite{orio2023dynamical}). 

This results in a synergy-vs-redundancy tradeoff \citep{luppi2024information,luppi2022synergistic,rosas2020operational,proca2024synergistic}. By explicitly separating these atoms, PID resolves classical ambiguities in complex systems—such as conflating ``integration-as-oneness'' (redundancy) with ``integration-as-cooperation'' (synergy)---clarifying distinct axes of integration, differentiation, and locality (Figure \ref{fig:triangle}b). Thus, even without engaging with the computational aspects of PID, the conceptual toolkit and vocabulary of redundant, unique, and synergistic information allow researchers to navigate important tradeoffs that are central for understanding complex systems.

% @Fernando: for space efficiency reasons, i think we cannot explain everything twice (once in the text and once in the caption). For now, i'm trying to offload everything to the figure caption.

\subsection{Emergence}\label{sec:primer_emergence}
% What makes PID useful to study emergence is that synergy provides a necessary condition: a system exhibits emergence only if it carries synergistic information about task-relevant outcomes. This criterion is falsifiable, quantitative, and it applies equally whether the ``parts'' are attention heads inside a transformer, agents in a multi-agent system, or a human and an AI collaborating on a task. 

% #########################
% Three tests
% #########################

Emergence is a rich topic that can be studied from various perspectives~\cite{gibb2019routledge,krakauer2023symmetry,carroll2024emergence,jensen2022complexity}, being the subject of long-standing ongoing debate. Complementing the rich philosophical literature~\cite{bedau2008emergence,clayton2006re}, PID provides a falsifiable, quantitative, and substrate-agnostic criterion for emergence: \emph{a system exhibits emergent properties only if it carries synergistic information about task-relevant outcomes}. 

How can one test if a system with multiple parts shows emergent properties? That is, how can one test whether the whole is more than the sum of its parts? Building on the idea of information decomposition, one can cast emergence as information about a system's temporal evolution (future states of the whole) that cannot be traced to the current state of its parts using time-delayed mutual information (TDMI) \citep[]{luppi2022synergistic}. This then gives a data-driven, quantifiable criterion amenable to empirical testing with falsifiable conjectures against specific null hypotheses \citep{riedl2026emergent}. Specifically, three tests of emergence have been proposed \citep{mediano2022greater}, which we describe next. 

% ###### Emergence capacity ######
\textbf{Emergence capacity.} 
This first test probes the ability of a system to host \textit{any} emergent variable. Specifically, for each pair of sources $(i,j)$ and timestep $t$, let the sources be $X_{i,t}$ and $X_{j,t}$, and define the bivariate target variable as the joint future state at $t+\ell$, $T_{ij,t+\ell} \equiv (X_{i,t+\ell}, X_{j,t+\ell})$, where $\ell$ is the integration timescale \citep{mediano2022integrated}. 
From the joint contingency table over $(X_{i,t}, X_{j,t}, T_{ij,t+\ell})$ we then compute the two-source PID using Eq.~\eqref{eq:pid}. A positive score of the pairwise dynamical synergy $\mathrm{Syn}_{ij} > 0$ indicates that the macro captures predictive information about the joint future not recoverable from any single component. A common approach is to compute this for all unordered pairs $(i,j)$ and then take the median as system-level synergy capacity \citep{riedl2026emergent}. This test is limited to detecting synergy of order $k=2$, but has the advantage that it does \textit{not} rely on the definition of a suitable whole-system macro signal beyond the states of the sources.  

% ###### direct test ######
\textbf{Unique information criterion.} 
This second test evaluates a specific macroscopic feature $V_t$ against the formal criteria for causal emergence \citep{mediano2022greater}: a feature $V_t$ is emergent if it possesses predictive power about the system’s future ($X_{t+\ell}$) that is unique with respect to the state of its individual parts.\footnote{More restrictive conditions include the $k$th-order unique information, as discussed in~\cite{rosas2020reconciling}.} This is quantified using the unique information atom from Eq.~\eqref{eq:pid}:
\begin{equation}\label{eq:direct_test}
    \text{Un}(V_t; X_{t+\ell} | X_t) > 0.
\end{equation}
This test determines if the macro-variable has access to information about the system's evolution that no subset of $k$ or fewer parts can access on their own. While this is the most theoretically rigorous way to identify an emergent feature, it is computationally demanding because it requires full knowledge of the statistics of the---potentially high-dimensional---microscopic dynamics.

% ###### practical criterion ######
\textbf{Practical criterion.} 
This third test concerns predicting a macro signal $V$ of the system by asking whether the macro contains predictive information beyond any individual part. Given the microstate $X_t = (X_{1,t}, \dots, X_{n,t})$, samples of the macro $V_t = f(X_t)$ are aligned as $(t, t+\ell)$. From this, one can compute the following score:
\begin{equation}
    \Psi_\ell(V) = I(V_t; V_{t+\ell}) - \sum_{k=1}^{n} I(X_{k,t}; V_{t+\ell}).
    \label{eq:practical_criterion}
\end{equation}
A positive value indicates that the macro's self-predictability exceeds what the sum of its parts can explain, indicating dynamical synergy. A benefit of this score is that it acts as a coarse, order-agnostic screen sensitive to multi-part synergy (i.e., synergy of any order $\geq 2$). However, it is penalized by redundancy across parts which can make the score negative even when higher-order synergy exists. In practical terms, the measure is often computed in a bias-corrected version against some permuted null model. Recent work has shown that Eq.~\eqref{eq:practical_criterion} can serve as a differentiable training objective, learning the macro variable $V$ rather than requiring it \textit{a priori} \citep{mcsharry2024learning}.

% 4 Formal Truths resulting from PID
% Maybe format as list if we have space
PID entails four architectural principles governing synergy in any system---circuit, agent team, or human-AI collaboration.
\begin{itemize}
    \item[(i)] \textbf{Differentiation requirement.} To generate synergy the system's components must be either functionally differentiated or modularized by context. Without differentiation---if the parts are perfectly synchronized, cloned, or otherwise highly coupled---they can only provide redundant information that cannot complement each other. Differentiation can be engineered (e.g., via personas \citep{riedl2026emergent}) or arise endogenously (e.g., via learned routing \citep{fulker2024spontaneous}).
    \item[(ii)] \textbf{Representational capacity.} Synergy provides functional advantages as it can encode more information than redundancy atoms for a given system size (representational or combinatorial capacity) \citep{rosas2020operational}. % trying to trim a bit; Synergistic information processing unlocks the flexibility to integrate and process various sources of information in different ways that is important for complex tasks.
    \item[(iii)] \textbf{Temporal emergence.} Emergent properties are not present in the instantaneous microstate alone, but arise diachronically through processes unfolding over time \citep{rosas2020reconciling}. % and/or bedau1997weak?
    \item[(iv)] \textbf{Fragility-vs-robustness tradeoff.} Synergy is associated with fragility, while redundancy is inherently robust~\cite{gutknecht2025shannon}. Because emergent synergy relies on precise structural heterogeneity, it is inherently unstable and highly sensitive to the specific input signals.
\end{itemize}

% \paragraph{Estimation.} 
% Estimation and data can be tricky for PID
% Binary, gaussian
% BC estimates
% Riedl et al develops two nulls; (explain row- and column-shuffle nulls)
% what else do we want / need here?

% #################################
% ## Three Worked Examples
% #################################
% \clearpage % TMP pagebreak
\section{Applications}\label{sec:applications}
This section provides three worked examples across different subfields, with applications on circuits, multi-agent systems, and human-AI interaction.

% #################################
% ## Ex1: Within LLM
% #################################
\subsection{Synergy within LLMs}\label{sec:ex1}
Recent work has shown that the human cortex is organized along a redundancy-to-synergy gradient: redundancy dominates in primary sensorimotor areas, whereas synergy concentrates in the central executive networks linked to higher cognition~\cite{luppi2024information}. Redundancy is largely conserved across primates~\cite{luppi2022synergistic} while synergistic interactions are selectively expanded in humans. % Furthermore, integrated information has been found to be disrupted under anesthesia~\cite{luppi2024synergistic}, and emergent dynamics are disrupted by traumatic brain injury~\cite{luppi2023reduced}. 
This informational ``separation of roles'' also occurs in artificial neural systems. 
For example, synergy increases as neural networks learn diverse tasks or modalities~\citep{proca2024synergistic,liang2023quantifying}. Tasks requiring input integration depend on synergistic neurons. 
Reservoir computers evolve toward and benefit from synergistic dynamics~\cite{tolle2026evolving}, suggesting that synergy is a selected computational substrate rather than a byproduct.

How is information processing organized within LLMs? \citet{urbina2026brain} treat attention heads and mixture-of-experts (MoE) modules as nodes, record each unit’s token-by-token activity as the L2 norm of its output, and apply a time-lagged PID ($\Phi$ID) with MMI redundancy to the resulting multivariate time series. Because a full decomposition over all units is intractable (Section \ref{sec:primer_synergy}), they compute bivariate $\Phi$ID for every pair of heads (or experts) which they aggregate into a per-unit synergy-redundancy rank. Synergy claims therefore reflect the density of pairwise, order-2 synergistic interactions rather than the full higher-order lattice.

\begin{figure}[b!]
    \centering
    \includegraphics[width=.80\textwidth]{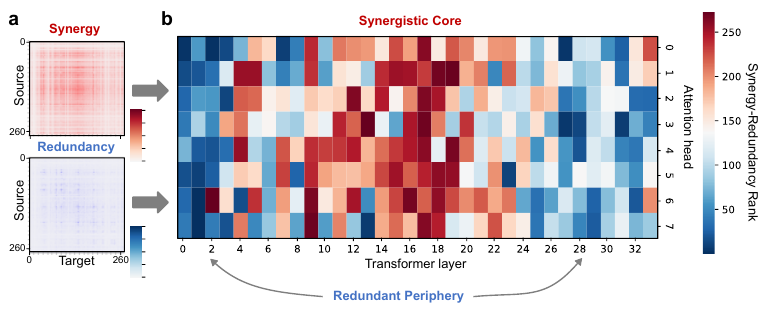}
    \vspace{-8pt} 
    \caption{
        LLMs exhibit a synergistic core surrounded by a redundant periphery. 
        \textbf{a)} Heatmaps of synergy and redundancy between pairs of attention heads in Gemma 3 4B. 
        \textbf{b)} Heatmap of the synergy-redundancy ranks in Gemma 3 4B.
        Figures adapted from \citet{urbina2026brain}.
    }
    \label{fig:ex1}
\end{figure}

Middle layers show synergistic processing while early and late layers exhibit a dominance of redundancy---a striking analog of the unimodal-to-transmodal gradient observed in the human brain. These results hold across multiple LLM families---including Llama 3, Qwen 3, and DeepSeek V2 Lite. Moreover, the synergistic core is learned, not hardwired: it is absent in randomly initialized networks and holds across both attention heads and routed experts, indicating it is a general property of trained transformers rather than an architectural artifact.
The synergistic core is causally necessary for capabilities that demand integration: ablating synergistic heads degrades complex task performance disproportionately compared to ablating redundant heads. Synergistic information thus collapses sharply under targeted perturbation while redundant codes degrade gracefully. The same PID decomposition has been applied to diffusion transformers, where synergistic prompt interactions specifically index compositional image generation \citep{dewan2024diffusion}, and to large networks via subsampling \citep{ehrlich2023a}. 

Going beyond diagnostics, RL-based fine-tuning of synergistic regions produces larger performance gains than fine-tuning redundant components, whereas under supervised fine-tuning, targeting synergistic regions confers no advantage. This suggests that synergistic representations require reinforcement-driven optimization to unlock.

% #################################
% ## Ex2: Multi-Agent Systems
% #################################
% \clearpage % TMP pagebreak
\subsection{Synergy Across Agents in Multi-Agent Systems}\label{sec:ex2}
When multi-agent systems outperform single agents \citep{wu2023autogen,chen2023agentverse,li2024more,tao2024magis}, authors often attribute this to ``greater-than-the-sum-of-its-parts'' effects \citep[][p.1]{chen2023agentverse}. To test if this emergence is genuine, \citet{riedl2026emergent} applied information decomposition and the emergence tests introduced above. Specifically, the paper developed a simple group task in which agents propose integers whose sum needs to match a hidden target number. Blind to group size and peers' guesses, agents receive only collective ``too high'' or ``too low'' feedback. The paper then reports results on experiments with 10 agents per group across different LLM models (e.g., GPT-4.1) with three randomized interventions. Groups in the control condition ($\cplain$) exhibit strong temporal synergy but little coordinated alignment across agents. In a persona condition ($\cpersona$), agents show stable identity-linked differentiation (e.g., some agents consistently guess lower numbers). A Theory of Mind ($\ctom$) condition instructs agents to ``think about what other agents might do.'' Here, agents additionally show goal-directed complementarity. 

\begin{figure}[t!]
    \centering
    \includegraphics[width=.80\textwidth]{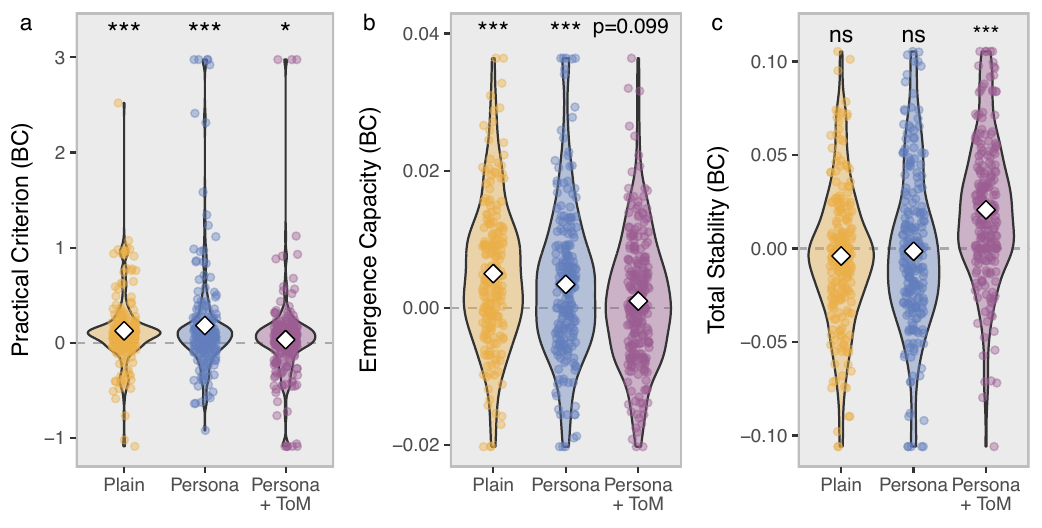}
    \vspace{-8pt} 
    \caption{
        Emergent synergy in multi-agent LLM systems.
        \textbf{a)} Practical emergence criterion (bias corrected). 
        \textbf{b)} Emergence capacity dynamical synergy (bias corrected).
        \textbf{c)} Total Stability (time-delayed mutual information of triplets).
        Data were Winsorized at the 1st and 99th percentiles for visual clarity. 
        Stars indicate significance level of Wilcoxon test. 
        Figures adapted from \citet{riedl2026emergent}.
        {\small Notes: *** $p < 0.001$; ** $p < 0.01$; * $p < 0.05$.}
    }
    \label{fig:ex2}
\end{figure}

Using group error (the difference between the group's guess and the target number) as macro signal $V$ and MMI as redundancy function, it is shown that both the emergence capacity criterion (Eq.~\eqref{eq:pid}) and the practical emergence criterion (Eq.~\eqref{eq:practical_criterion}) are significantly above zero, confirming dynamical emergence (Figure \ref{fig:ex2}a-b). 
To probe how this emergence is maintained, they compute Total Stability---defined as the triplet-wise mutual information normalized by macro-signal entropy. Total Stability is indistinguishable from zero in the $\cplain$ and $\cpersona$ conditions but sharply elevated in the $\ctom$ condition  ($p$-value = $2.9 \times 10^{-14}$; Figure \ref{fig:ex2}c). This suggests the ToM intervention to ``think about other agents'' acts as a control parameter, steering the multi-agent system from a chaotic regime into a deep basin of attraction where collective behavior stabilizes and information processing capacity spikes. These results are corroborated through various robustness tests (full vs.~early rounds only, raw measures, different bias correction approaches, a functional null to account for autocorrelation, and two different redundancy measures).

Finally, results show that neither synergy nor redundancy alone predicts success, while their interaction does ($\beta = 0.24$; $p = 0.014$): pairwise alignment to the macro signal supplies goal alignment (redundancy), while identity-linked differentiation in the $\cpersona$ and $\ctom$ conditions supplies complementary roles (synergy). Performance improves only when both are present, illustrating that multi-agent coordination requires balancing integration with differentiation across the PID lattice.
Human groups have been shown to exhibit the same signature: collective intelligence requires both alignment and complementarity \citep{riedl2017teams,riedl2021quantifying}.

% #################################
% ## Ex3: Human-AI Synergy
% #################################

% #################################
% ## Ex3: Human-AI Synergy
% #################################
\subsection{Human-AI Synergy}\label{sec:ex3}
Studies of human-AI interaction typically measure performance uplift by comparing between a ``human-only'' and a ``human-and-AI'' condition \citep[e.g.,][]{doshi2024generative,dell2026navigating,brynjolfsson2025generative,noy2023experimental}. Those comparisons do not distinguish whether gains are additive, substitutive, or non-linearly synergistic and presuppose a single-agent baseline, which excludes tasks no single agent can attempt.\footnote{Tasks with no single-agent baseline can still be analyzed, but need randomized composition (ideally repeated; \citep{weidmann2021team}) and off-task skill ratings as the parts’ marginals \citep{riedl2017teams}.} \citet{riedl2026quantifying} studied human-AI collaboration using item response theory (IRT), isolating latent AI ability, human solo ability, and human collaborative ability. While IRT latent traits model the cognitive data-generating process rather than yielding information-theoretic quantities, we can map this behavioral decomposition onto the conceptual lattice of PID. 

Considering the contribution of the human ($H$) and the AI ($A$) to the behavioral outcome ($Y$), the joint information decomposes exhaustively into four non-negative atoms based on Eq.~\eqref{eq:pid}:
\[
    I(H, A;\, Y) = \mathrm{Un}(H) + \mathrm{Un}(A) + \mathrm{Red}(H, A;\, Y) + \mathrm{Syn}(H, A;\, Y).
\]
Because the emergence criterion (Section \ref{sec:primer_emergence}) requires only $\mathrm{Syn}>0$, one does not need to isolate the synergistic atom directly: it suffices to \textit{bound} the other three and show that a positive residual survives. The behavioral tests in Ref.~\citep{riedl2026quantifying} provide a sufficient basis to bound each of them.\footnote{The same study also estimates a formal PID and finds a credible $\mathrm{Syn}(H, A; Y) > 0$.}

\textit{The residual is not unique human information ($\mathrm{Un}(H)$).} One potential concern is that the joint outcome is the human's solo knowledge re-expressed, loading the residual onto $\mathrm{Un}(H)$. This is ruled out because the human's solo ability ($\theta$) and collaborative ability ($\kappa$) are statistically distinct latent traits (using Expected Log-Predictive Density $\Delta$ELPD = 50.9, SE = 10.2; $\theta \neq \kappa$). While distinct traits could yield identical information, discerning evidence comes from the AI-unsolvable regime: users improve on items neither source could resolve alone. Thus, the distinct collaborative process ($\kappa$) generates novel information absent from $H$'s marginal.

\textit{The residual is not unique AI information ($\mathrm{Un}(A)$).} Another potential concern is delegation: the AI supplies the correct answer and the user adopts it, loading the residual onto $\mathrm{Un}(A)$. One can bound this channel by forcing $\mathrm{Un}(A)$ to zero. The fundamental properties of PID dictate that $\mathrm{Un}(A) \leq I(A;\,Y)$. By splitting the joint condition and observing items the AI could not solve alone, we restrict the analysis to a regime where $I(A;\,Y) \approx 0$, thereby shutting off $\mathrm{Un}(A)$. Users still improve significantly over their solo baseline on these specific items ($\beta = 0.80$ [0.59, 1.01]). Furthermore, user-level collaborative slopes on AI-solvable and AI-unsolvable items are highly correlated ($\rho = 0.80$ [0.53, 0.96]), contradicting the regime-specific pattern delegation predicts. Improvement that persists where $\mathrm{Un}(A)$ is mathematically starved cannot be AI-unique information.

\textit{Redundancy is starved on the hard-item regime.} Under MMI redundancy \citep{barrett2015exploration} (the same redundancy measure used in Sections \ref{sec:ex1} \& \ref{sec:ex2}) $\mathrm{Red}(H,A;\,Y) = \min(I(H;\,Y), I(A;\,Y))$, so as questions become harder for the human alone, $I(H;\,Y) \to 0$ drives $\mathrm{Red} \to 0$ with it. We do not bound the aggregate atom; we isolate a regime where it is structurally near zero. The behavioral data track this gradient: users switch from an incorrect solo answer to a correct joint one on 17-19\% of the easiest items but 31-38\% of the hardest, and co-success is highest in moral reasoning (where humans often already hold the answer) and lowest in physics. On hard items---where $\mathrm{Red} \approx 0$ and, when the AI alone cannot solve them, $\mathrm{Un}(A) \approx 0$---the surviving residual is almost pure $\mathrm{Syn}$ (plus any elicited human knowledge, itself synergistic by the argument below). Because MMI upper-bounds every valid redundancy measure and $\mathrm{Red} \approx 0$ in this regime, this $\mathrm{Syn}$ estimate does not depend on the choice of redundancy measure.

\textit{The residual carries the structural fingerprint of synergy.} What remains is not merely a leftover: it behaves strictly as an interaction-dependent atom should. Human ToM predicts collaborative but not solo ability, and when ToM is zero, access to AI provides no benefit ($\beta = -0.07$ [-0.84, 0.70]). Unique and redundant information do not require joint processing to exist; only synergy is contingent on the effective cognitive engagement of both sources. The vanishing benefit of the joint condition when ToM is zero is thus a positive behavioral signature of $\mathrm{Syn}$.

Finally, one might suggest that the AI is simply eliciting the human's own unexpressed, latent knowledge, rather than creating novel information. Under PID, this is not a caveat, but rather the very definition of synergy. If a human possesses latent knowledge that contains information about $Y$, but that information only manifests in the observable outcome when triggered by the AI's interaction, the information exists strictly in the joint state $(H, A)$. It cannot be found in the marginal distribution of the human's solo responses. Information that is generated exclusively through the interaction of sources is, axiomatically, synergistic.
% (the same argument holds in the opposite direction where the human elicits information from the AI through their specific way of promoting)

\textit{The synergistic atom is generated by unilateral adaptation.} Although PID synergy is non-additive \textit{information} by definition, the mechanism generating it here is additive at the latent level: collaborative residuals are under-dispersed ($\phi = 0.658$), and user-by-difficulty interactions add no predictive value ($\Delta\text{ELPD} = -1.7$; 80\% power to detect non-linearity as small as 16\% of the additive effect). Synergy is thus extracted through a stable, general channel (the human's collaborative ability $\kappa$, expressed as conditional delegation, scaffolding, interrogation, and calibrated reliance on AI output) rather than through item-specific feedback loops. This is the signature of an asymmetric system in which only the human adapts. Reciprocal adaptation by the AI would add trial-level interaction variance ($\phi > 1$), though with a single AI partner that dyadic term is not identifiable \citep{kenny1984social}. The resulting information is genuinely synergistic, but its generation is currently bounded by one-sided adaptation.

\begin{figure}
    \centering
    \includegraphics[width=.95\textwidth]{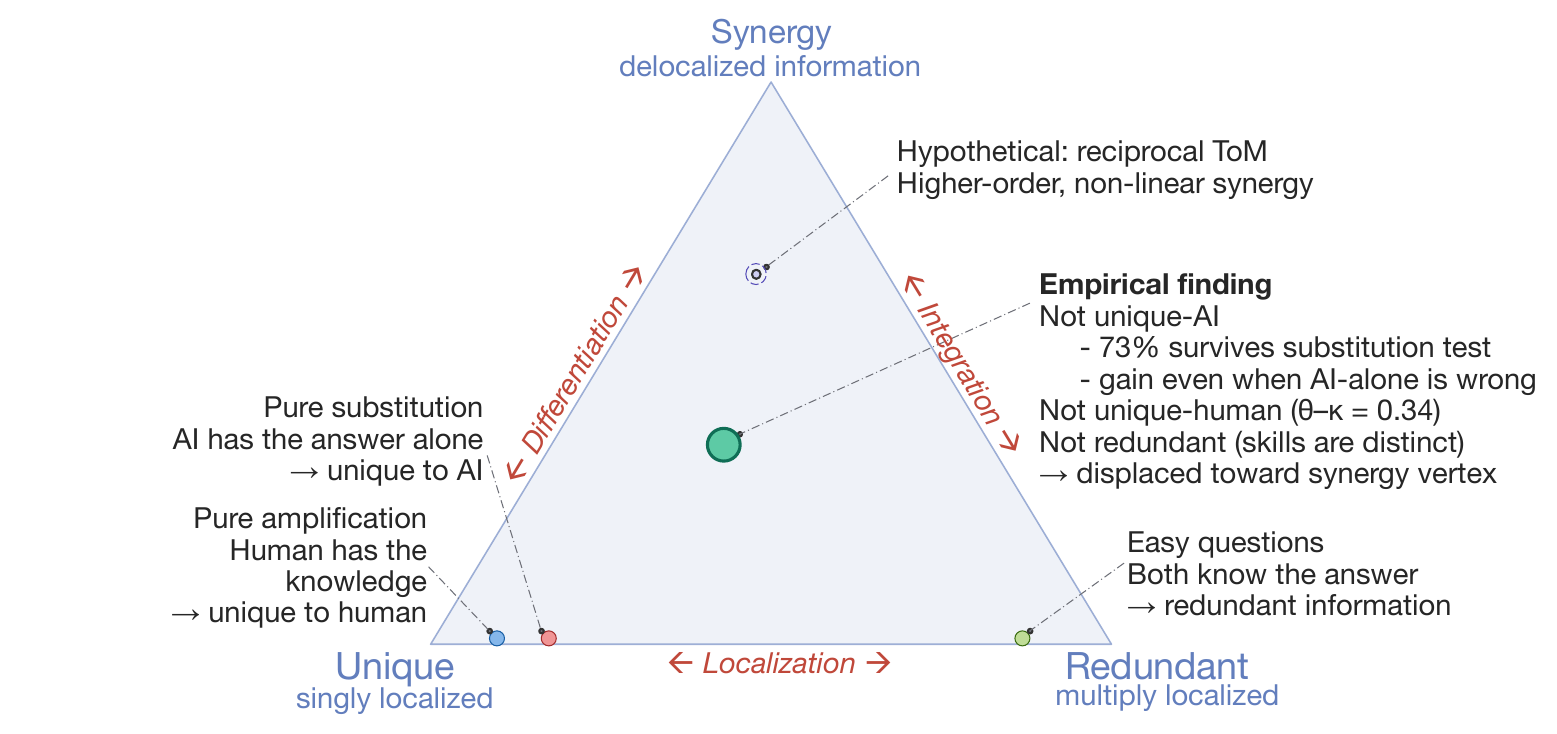}
    \vspace{-8pt} 
    \caption{
        Viewing human-AI synergy through an information decomposition lens unifies existing theories about substitution, linear and non-linear complementarity. 
        % Variance finding: individually individuals are spread out more than with AI.
    }
    \label{fig:ex3}
\end{figure}

Taken together, these tests leave a positive residual that none of the other three atoms can absorb, which is a lower bound on $\mathrm{Syn}(H,A;Y)$ and exactly what the emergence criterion requires, obtained from evidence never collected with PID in mind  (Figure~\ref{fig:ex3}).

% #################################
% ## Summary after the three worked examples
% #################################
\subsection{Scope of these claims} 
The choice of parts, target, timescale, and redundancy measure is not given by the theory. In an LLM, ``part'' could mean neuron, attention head, expert, or layer. Claims about synergy and emergence are bounded by those modeling choices which should be founded on theoretical arguments and supplemented by robustness tests. The synergistic-cores work, for example, finds signs of synergy across different granularities (attention heads \& experts in MoE); the multi-agent application explores robustness across three different targets (it uses joint future state of agent pairs, group error, and first principal component as macro signal) and redundancy measures ($I_{min}$ and $I_{\mathrm{MMI}}$). Claims also differ in how far they climb an evidential ladder: detection, localization, causal identification, control, and prediction (Table \ref{tab:evidence_ladder}). We summarize practical guidelines for applying our framework in Table \ref{tab:practical-guidelines} and reporting its results in Box \ref{box:reporting}. We provide a quick start guide with easy-to-use code and illustrations at \url{https://github.com/riedlc/emergence}.

% #################################
% ## DISCUSSION
% #################################
% \clearpage
\section{Discussion}\label{sec:discussion}
In this paper we have argued that Partial Information Decomposition provides a unifying lens to study synergy across scales and substrates of AI systems---from internal components to multi-agent ensembles and human-AI interaction. While current synergy metrics vary by domain, they implicitly target a shared structure: information distribution across unique, redundant, and interaction-dependent components. Grounding metrics in this framework allows us to compare abstraction levels, clarify definitions of emergence, and advance a cumulative science of collective intelligence across scales and substrates. 
This lens unifies existing practices as design levers for navigating the synergy-redundancy spectrum, which we summarize in Table \ref{tab:engineering_view}. Additionally, this perspective reveals hidden cross-layer connections to guide future research which we summarize in the five conjectures below (with testable hypotheses in Table \ref{tab:testable}).

% I.e., there is a specific mechanistic chain: PID -> gradient structure -> loss landscape flatness -> delayed generalization
\paragraph{Conjecture 1: Grokking as redundancy-to-synergy transition.} Grokking, the abrupt onset of generalization long after memorization \citep{power2022grokking,nanda2023progress}, admits a natural PID account. While unique and redundant information provide immediate, dense gradients during early training, synergistic targets require higher-order coordination and yield negligible marginal mutual information \citep{abbe2023sgd,barak2022hidden}, leaving per-weight gradients flat. Optimization is therefore initially biased toward redundant representations, and synergistic representations emerge only when the optimizer  discovers a coordinated weight configuration that unlocks a non-zero synergistic gradient \citep{cullen2026grokking}. We predict grokking should correspond to a measurable phase transition in the PID spectrum: redundancy-dominated during memorization, shifting abruptly to synergy at the generalization jump. We also predict that regularizers suppressing higher-order dependencies should delay or prevent generalization while leaving memorization intact. This conjecture is partially supported by the synergy--redundancy shift documented by \citet{clauw2024information}, and  
is closely related with work on 
\emph{leap complexity}~\cite{abbe2023sgd} showing that complex Boolean functions (maximally synergistic parities characterized by high-degree Fourier coefficients) are learned via saddle-to-saddle dynamics~\cite{barak2022hidden}.

\paragraph{Conjecture 2: Jagged intelligence as a factored-representation boundary.}
LLMs pass the bar exam yet fail basic logic puzzles---a pattern termed ``jagged intelligence'' \citep{dell2026navigating,song2026large}. However, existing accounts lack a structural criterion for predicting \textit{which} tasks will fall on which side of the frontier. The factored-representation bias described in Section~\ref{sec:ex1} suggests one: models should fail where tasks require irreducible cross-factor integration. We suggest task difficulty tracks not human-perceived difficulty but \textit{synergistic depth}: the degree to which the correct answer depends on jointly considering independently informative factors. Consider, for instance, the viral LLM failure ``should you walk or drive to the car wash 50 meters away?''.\footnote{\url{https://opper.ai/blog/car-wash-test}} Distance suggests walking and washing suggests bringing the object but only integrating the two reveals that the car \textit{is} the object. The ablation test in Section~\ref{sec:ex1} suggests a causal mechanism: middle layers, home to the synergistic core, appear prunable on benchmarks that do not probe cross-factor integration, but become indispensable once the task demands it. Independent evidence agrees: synergistic information is harder for models to capture than redundant information \citep{liang2023quantifying}. Furthermore, causal-abstraction work on belief tracking \citep{prakash2026language} finds that models default to factored representations of per-agent beliefs, and recruit dedicated integration heads only when cross-agent dependencies demand it.

\paragraph{Conjecture 3: Test-time compute as synergy unrolling.} Recent information-theoretic analysis of chain-of-thought (CoT) reasoning formalizes each step as a conditional information gain, showing that correct steps monotonically increase information about the target while incorrect steps yield zero gain \citep{ton2025understanding}. The PID lens suggests a specific structural interpretation: because a single forward pass has fixed functional depth, tasks whose solutions require high synergistic depth cannot be resolved in one pass. CoT acts as an algorithmic compressor of per-step synergistic load, externalizing partial states into the context window and converting within-pass synergistic integration into sequential redundant lookbacks across tokens \citep{prakash2026language}. This yields two predictions: (i) a reasoning model's internal representations should carry lower per-pass synergistic load than a standard LLM's, because synergy has been offloaded into the token stream; and (ii) tasks with irreducibly high synergistic depth---those resisting decomposition into sequential sub-problems---should show diminishing returns from additional test-time compute.

\paragraph{Conjecture 4: Multi-turn degradation and path-dependent synergy collapse.} LLMs degrade when instructions are split across turns, collapsing at two turns rather than declining gradually \citep{laban2026llms}, even though the full conversation remains in context. Synergistic information requires that all sources are not only present but jointly processed. Anything that degrades joint access can destroy synergistic channels without removing any information. Examples include salience shifts from reordering, and intermediate responses computed from marginals that commit the context  prematurely. We predict that multi-turn degradation, input-order sensitivity, and prompt-sequence effects in multi-agent settings share one cause: disrupted access to synergistic information. All three should scale with a task's synergistic depth and leave redundancy-dominated tasks unaffected. \citet{laban2026llms} observe both the cliff-at-two-turns pattern and an episodic-task exception consistent with this account.

\paragraph{Conjecture 5: Theory of mind as a cross-level control parameter.} Across all three levels, a structurally analogous mechanism shifts systems from factored to synergistic processing. In LLMs, sentences that introduce belief discrepancies activate dedicated synergistic integration heads \citep{prakash2026language}; in multi-agent groups, ToM prompting shifts coordination toward synergy dominance \citep{riedl2026emergent}; and in human-AI teams, human ToM predicts collaborative but not individual ability \citep{riedl2026quantifying}. In each case, one component accesses another's internal state, creating conditions for irreducible joint information. This yields a cross-level prediction that neither community would make alone: high-ToM human inputs should differentially activate the model's synergistic core, a measurable circuit-level signature of a behavioral interaction variable.

% #################################
% ## ALTERNATIVE VIEWS
% ## Required section for "perspective" pieces
% #################################
\section{Alternative Views}\label{sec:alternative_views}

\paragraph{PID is the wrong foundation.} One might argue that simpler measures---like mutual information---are useful without the overhead of full information decomposition. We agree, and do not advocate replacing established methods. However, because simpler metrics conflate fundamentally different phenomena (e.g., synergy vs.~redundancy), researchers must rely on complementary tests and triangulation to bound estimates and rule out alternative accounts. PID's value is not as a mandatory measurement tool, but as a theoretical scaffold specifying what these simpler empirical tests capture.

\paragraph{O-information or other measures would work just as well.} Formal PID is computationally demanding and data-hungry. Alternatives like O-information or integrated information capture related higher-order structures as complementary, tractable summaries. Our argument is not that PID is the only useful metric, but that its conceptual vocabulary---distinguishing synergy from redundancy---is necessary to make emergence claims precise, whichever estimator tests them.

\paragraph{Other frameworks already study higher-order interactions.}  Tensor networks \citep{kolda2009tensor} and higher-order networks (hypergraphs, simplicial complexes; \citep{battiston2020networks}) describe the structure over which interactions occur; mechanistic null models such as the game-theoretic block model \citep{riedl2026gtbm} test whether that structure is reducible to pairwise interaction, but cannot type the residual they leave; causal abstraction \citep{geiger2025causal} and related accounts of emergence \citep{hoel2013quantifying,barnett2023dynamical,rosas2024software,rosas2025symmetries} identify macroscopic variables via scale separation. 
PID does not compete but complement these approaches. 
None of them distinguishes the informational type of an interaction relative to a target, which is what an emergence-as-irreducibility claim turns on. 
PID supplies that distinction. 
It therefore does not compete with these frameworks so much as label the atoms they organize, transport, or coarse-grain.

\paragraph{These are three different phenomena, not one.} A natural objection is that grouping transformers, multi-agent systems, and human-AI teams obscures their vastly different mechanisms. We acknowledge these differences in scale and substrate. However, while their physical forms diverge, their functional and information-processing architectures share similar structures. Information decomposition allows us to translate findings across scales and formulate testable, cross-layer predictions that siloed fields would miss. Ultimately, this is an empirical question: the success of these hypotheses will dictate whether this framework is based on a mere analogy or on shared computational reality.

% We don't have the space, but this would be a great point to make too
% \paragraph{Performance is all I care about.} A natural objection is that a practitioner who only wants the best-performing system has no use for decomposing where its performance comes from. But performance is a level, and nearly every decision a practitioner makes — scale the system, prune it, deploy it on harder inputs, intervene on its design — depends on how that level responds to change, which is determined by the composition of the gain rather than its size. Redundant gains saturate under scaling while synergistic gains grow combinatorially (§2.1); synergistic performance collapses under ablation while redundant performance degrades gracefully (§3.1); human-AI gains that are unique to the AI vanish precisely on items the AI cannot solve, while synergistic gains persist (§3.3); and the lever that raises performance differs by atom (Table 1). Each of these decisions is a counterfactual, and each counterfactual experiment estimates a sum of PID atoms (Section 2). The decomposition is what those experiments are projections of; estimating it directly is what lets their answers transfer.

% #################################
% ## Acknowledgments
% #################################
\begin{ack}
C.R. received no funding for this work.
F.E.R. has been supported by the UK AI Security Institute and Coefficient Giving. 
The authors declare no competing interests.
\end{ack}

% #################################
% ## REFERENCES
% #################################
\bibliographystyle{plainnat}
\bibliography{references,references_software}

%%%%%%%%%%%%%%%%%%%%%%%%%%%%%%%%%%%%%%%%%%%%%%%%%%%%%%%%%%%%
\clearpage
\appendix
\renewcommand{\thetable}{S\arabic{table}}
\renewcommand{\thefigure}{S\arabic{figure}}
\setcounter{table}{0}
\setcounter{figure}{0}

\section{Supplementary Material}

\subsection{Technical introduction to Partial Information Decomposition}
\label{sup:technical_details}

Let us provide here a more technical explanation of the foundations of Partial Information Decomposition~\cite{williams2010nonnegative}. 
The general decomposition of the information provided by variables $\bm X=(X_1,\dots,X_n)$ about another variable $Y$ is built following these steps.
\begin{itemize}
    \item[1)] Define to be a generalized ``source'' of information all subsets of variables. Generalized sources can be indexed with numbers within round brackets. For example, the source constituted by variables $X_1$, $X_2$, and $X_4$ is denoted by $(124)$. 
    \item[2)] Establish a method to quantify the redundancy between generalized sources, denoted by $I_\cap$. This should capture the amount of information provided by multiple groups of variables; for instance, the redundancy between sources $(12)$ and $(145)$ is captured by the quantity $I^{(12)(145)}_\cap(\bm X;Y)$. As a consistency consideration, the redundancy of a source with a single group of variables is required to be equal to the corresponding mutual information, so that $I^{(13)}_\cap(\bm X;Y)= I(X_1,X_3;Y)$.
    \item[3)] Assume that the redundancy between $m$ generalized sources $S_1,\dots,S_m$ doesn't change if one removes sources that are contained within other sources. For example, $I^{(12)(34)(2)}(\bm X;Y)=I^{(12)(34)}(\bm X;Y)$. This implies that one should focus on the redundancy between collections of generalized sources that form ``antichains'', where no generalized source is fully contained by another. This also suggests that the redundancy should monotonically follow the following partial relationship: 
    % \begin{equation}
    %    \bm \alpha \leq \bm \beta \quad \iff \quad \forall A\in\bm\alpha,\exists B\in\bm\beta, A\subseteq B. \label{eq:redundancy_ordering}
    %\end{equation}
    % @Fernando: I believe one ordering was swapped
    \begin{equation}
        \bm \alpha \leq \bm \beta \quad \iff \quad \forall B\in\bm\beta,\exists A\in\bm\alpha, A\subseteq B. \label{eq:redundancy_ordering}
    \end{equation}
    The set of all antichains of $n$ source variables is denoted by $A_n$, and together with the partial order $\leq$ constitute the so-called \emph{redundancy lattice} $\mathcal{R}_n$. 
    \item[4)] Finally, the PID atoms are obtained by taking the M\"obius inversion over the redundancy lattice; i.e. are given by the terms $I_\partial^{\bm \beta}(\bm X; Y)$ that satisfy the following property:
\begin{align}
    I_\cap^{\bm \alpha}(\bm X; Y) := \sum_{\substack{\bm \beta\in A_n\\\bm\beta\leq \bm\alpha}} I_\partial^{\bm\beta}(\bm X; Y). \label{eq:PID_sum}
\end{align}
    Informally, this can be thought of as the ``derivative'' of $I_\cap^{\bm \beta}(\bm X; Y)$ on $\mathcal{R}_n$.
\end{itemize}

Using the PID atoms, the total information $I(\bm X;Y)$ can be decomposed as
    \begin{equation}
    I(\bm X;Y) = \sum_{\bm \alpha\in A_n} I_\partial^{\bm\alpha}(\bm X;Y).
    \end{equation}

The formulation above leaves the redundancy function $I_\cap$ unspecified; the minimum-mutual-information choice in Eq.~\eqref{eq:MMI} is one such function. MMI has a property worth flagging: it can give non-zero redundancy for uncorrelated sources. This feature motivates the distinction between \textit{source redundancy}, corresponding to correlations between sources that are informative about the target, and \textit{mechanistic redundancy}, which is redundancy associated to the structure of the mechanism (the channel) mapping sources to target and not from correlation among sources~\cite{harder2013bivariate,ince2017partial}.

\subsection{Example of algorithm to calculate synergy}

This appendix introduces a specific way to operationalize synergy, and discusses the algorithm used for its calculation, summarizing results first introduced in Ref.~\cite{rassouli2019data}. 

\subsubsection{Synergy and data privacy}

To outline the core idea, consider $\bm X=(X_1,\dots,X_n)$ to be $n$ discrete source variables and $Y$ a target variable following a distribution $p_{\bm X,Y}$. For a subsystem $\alpha\subset[n]:=\{1,\dots,n\}$, write $\bm X^{\alpha}$ for
the corresponding tuple of variables, and let a \emph{constraint set}
$\bm\alpha=\{\alpha_1,\dots,\alpha_L\}$ collect subsystems that we wish to
keep private. 
A channel $p_{V\mid\bm X}$ is called $\bm\alpha$-\emph{synergistic} if
its output $V$ is independent of every subsystem in $\bm\alpha$,
$V\perp\bm X^{\alpha_i}$ for all $i$, so that $I(V;\bm X^{\alpha_i})=0$ for all $i$. The set of such channels is denoted $\mathcal{C}(\bm X;\bm\alpha)$. Intuitively, a variable $V$ obtained via an $\bm\alpha$-synergistic channel may carry information about the whole $\bm X$ while betraying nothing about any private components. 

Building on these ideas, the $\bm\alpha$-synergy is the information about $Y$ that any such $\bm\alpha$-private observable $V$ can convey,
\begin{equation}
  S^{\bm\alpha}(\bm X\to Y)
  \;:=\;
  \sup_{\substack{p_{V\mid\bm X}\in\mathcal{C}(\bm X;\bm\alpha)\\ V-\bm X-Y}}
  I(V;Y).
  \label{eq:syndef}
\end{equation}
This has a direct operational reading: $S^{\bm\alpha}(\bm X\to Y)$ is the amount
of information about $Y$ that can be \emph{disclosed} from $\bm X$ while keeping
$\bm X^{\alpha_1},\dots,\bm X^{\alpha_L}$ perfectly private. Two boundary cases
anchor the intuition: with no constraints, $S^{\varnothing}(\bm X\to Y)=I(\bm X;Y)$,
whereas protecting every individual source, $\bm\alpha=\{\{1\},\dots,\{n\}\}$,
isolates the information that survives only in the joint behavior of the
sources.

\subsubsection{The algorithm}
\label{app:syndisc-alg}

The supremum in~\eqref{eq:syndef} is always attained, and the optimal channel
is the solution of a linear program~\cite[Thm.~1]{rassouli2019data}. The key structural property to build such an algorithm is that a channel lies in
$\mathcal{C}(\bm X;\bm\alpha)$ if and only if each of its reverse conditionals
$p_{\bm X\mid V=v}$ preserves the protected marginals of $p_{\bm X}$. Collect
those marginals into a binary \emph{masking matrix} $\Pmask$, formed by
stacking, for each $\alpha_i\in\bm\alpha$, the (deterministic) map that sends a
distribution on $\bm X$ to the marginal of $\bm X^{\alpha_i}$. The feasible
reverse conditionals are then exactly the distributions in the polytope
\begin{equation}
  \mathcal{Q}_{\bm\alpha}
  \;=\;
  \bigl\{\, q\in\simplex\!\left(\supp p_{\bm X}\right)
  \;:\; \Pmask\, q = \Pmask\, p_{\bm X} \,\bigr\},
  \label{eq:polytope}
\end{equation}
i.e.\ the distributions that leave every protected marginal untouched. Because
$I(V;Y)=H(Y)-\sum_v p(v)\,H(Y\mid V=v)$ is convex in the channel, its maximum
over the feasible set is attained at the extreme points of
$\mathcal{Q}_{\bm\alpha}$. Enumerating those vertices $\{q_1,\dots,q_M\}$ fixes,
for each $j$, an induced target conditional $r_j=p_{Y\mid V=j}$ with entropy
$h_j=H(r_j)$; all that remains is to choose mixing weights $w_j\ge 0$ that
reconstruct the prior, $\sum_j w_j q_j = p_{\bm X}$, so as to minimize the
expected residual uncertainty $\sum_j w_j h_j$. The whole procedure is summarized as Algorithm~\ref{alg:syndisc}.

\begin{algorithm}[h]
\caption{Synergistic disclosure $S^{\bm\alpha}(\bm X\to Y)$ of a discrete system}
\label{alg:syndisc}
\begin{algorithmic}[1]
\Function{SynDisc}{$p_{\bm X,Y}$, $\bm\alpha=\{\alpha_1,\dots,\alpha_L\}$}
  \State $p_{\bm X}\gets\sum_{y}p_{\bm X,Y}(\cdot,y)$
         %\Comment{source prior on $\supp p_{\bm X}$}
  \State $p_{Y\mid\bm X}\gets p_{\bm X,Y}\,/\,p_{\bm X}$
  \State $H_Y\gets H\!\bigl(\textstyle\sum_{\bm X}p_{Y\mid\bm X}(\cdot\mid\bm X)\,p_{\bm X}(\bm X)\bigr)$
  \State build masking matrix $\Pmask$%: stack, for each $\alpha_i$,
         %the $0/1$ map $\bm X\mapsto\bm X^{\alpha_i}$
         %\Comment{leak-free marginals; App.~A of~\cite{rosas2020}}
  \State $\mathcal{Q}\gets\vertset\bigl\{\,q\in\simplex(\supp p_{\bm X}):
         \Pmask\,q=\Pmask\,p_{\bm X}\,\bigr\}$
         %\Comment{polytope~\eqref{eq:polytope}; Lemma~4}
  \For{each vertex $q_j\in\mathcal{Q}$}
    \State $r_j\gets\sum_{\bm X}p_{Y\mid\bm X}(\cdot\mid\bm X)\,q_j(\bm X)$
    \State $h_j\gets H(r_j)$
          % \Comment{induced $p(Y\mid V{=}j)$ and its entropy}
  \EndFor
  \State solve the linear program
  \Statex\hspace{2.4em}
         $\displaystyle w^\star=\argmin_{w\ge 0}\ \sum_j w_j\,h_j
         \quad\text{s.t.}\quad \sum_j w_j\,q_j=p_{\bm X}$
         %\Comment{$\sum_j w_j=1$ is implied}
  \State \Return $H_Y-\sum_j w^\star_j\,h_j$
         %\Comment{$=\max_{p_{V\mid\bm X}}I(V;Y)=S^{\bm\alpha}(\bm X\to Y)$}
\EndFunction
\end{algorithmic}
\end{algorithm}

%\begin{algorithm}[h]
%\caption{Bivariate disclosure decomposition via M\"obius inversion}
%\label{alg:pid2}
%\begin{algorithmic}[1]
%\Function{Decompose2}{$p_{X_1,X_2,Y}$}
%  \State $I(X_1,X_2;Y)$
%         \Comment{$=S^{\varnothing}$, the lattice bottom}
%  \State $S_{12}\gets\Call{SynDisc}{p,\{\{1\},\{2\}\}}$
%         \Comment{protect both sources: the synergy node}
%  \State $S_{1}\gets\Call{SynDisc}{p,\{\{1\}\}}$;\quad
%         $S_{2}\gets\Call{SynDisc}{p,\{\{2\}\}}$
%  \State $\Syn\gets S_{12}$
%         \Comment{$S^{\{1\}\{2\}}_{\partial}$}
%  \State $\Un_1\gets S_{2}-S_{12}$;\quad
%         $\Un_2\gets S_{1}-S_{12}$
%         \Comment{index flip: $\Un_1$ from the node protecting $X_2$}
%  \State $\Red\gets I-S_{1}-S_{2}+S_{12}$
%         \Comment{$S^{\varnothing}_{\partial}$}
%  \State \Return $(\Red,\Un_1,\Un_2,\Syn)$
%         \Comment{sums exactly to $I(X_1,X_2;Y)$}
%\EndFunction
%\end{algorithmic}
%\end{algorithm}
%

\subsubsection{Example: the {\upshape AND} gate}

To illustrate the algorithm, consider independent fair inputs
$X_1,X_2\in\{0,1\}$ and $Y=X_1\wedge X_2$. Then $p_{\bm X}$ is uniform on the
four states $\{00,01,10,11\}$, the target is deterministic given the inputs,
and $I(\bm X;Y)=H(Y)=H\!\left(\tfrac14\right)=0.8113$ bits.

To calculate synergy (i.e. $\bm\alpha=\{\{1\},\{2\}\}$.), we need to consider channels that keep private both marginals. This means that every feasible $q\in\mathcal{Q}_{\bm\alpha}$
must keep $\Pr(X_1{=}1)=\Pr(X_2{=}1)=\tfrac12$. On the support of $p_{\bm X}$
this leaves a one-parameter family,
$q=(t,\,\tfrac12-t,\,\tfrac12-t,\,t)$ over $(00,01,10,11)$ for
$t\in[0,\tfrac12]$---a line segment whose two vertices are
\begin{equation*}
  q_A=\bigl(0,\tfrac12,\tfrac12,0\bigr)\ \text{(anti-diagonal)},
  \qquad
  q_B=\bigl(\tfrac12,0,0,\tfrac12\bigr)\ \text{(diagonal)}.
\end{equation*}
Their induced target conditionals are $r_A=p(Y\mid V{=}A)=(1,0)$, since both
$01$ and $10$ map to $Y=0$, giving $h_A=0$; and
$r_B=p(Y\mid V{=}B)=(\tfrac12,\tfrac12)$, since $00\mapsto0$ and $11\mapsto1$,
giving $h_B=1$. The prior is the segment midpoint, so the reconstruction
constraint $w_A q_A+w_B q_B=p_{\bm X}$ pins the weights to
$w_A=w_B=\tfrac12$---here the linear program is forced. Hence
\begin{equation*}
  S^{\{1\}\{2\}}(\bm X\to Y)
  = H(Y)-\bigl(\tfrac12 h_A+\tfrac12 h_B\bigr)
  = 0.8113-0.5
  = 0.3113\ \text{bits}.
\end{equation*}
Operationally, disclosing the parity $X_1\oplus X_2$ narrows $Y$ (it excludes
the $11$ state whenever the inputs differ) without revealing either input,
and $0.3113$ bits is the most any leak-free observable can achieve.

\subsection{Practical considerations and design levers}
\label{app:syndisc-practice}

The example above exposes both the appeal and the cost of the approach. The appeal is that every atom is the value of a transparent optimization with an operational meaning; there is nothing to postulate. The cost is concentrated in one step---vertex enumeration of the polytope~\eqref{eq:polytope}---whose size is governed by $\dim\mathcal{Q}_{\bm\alpha}=|\supp p_{\bm X}|-\operatorname{rank}\Pmask$. For $n=2$ with both marginals protected this polytope is a line segment (two vertices) and the linear program is trivial, as we saw; but $|\supp p_{\bm X}|$ grows exponentially in the number of sources, so exact synergistic disclosure is a small-$n$ instrument. The \emph{backbone} decomposition mitigates the combinatorial explosion of the \emph{lattice}---its number of nodes grows only linearly in $n$---but each backbone term still solves a program over the joint support, so it does not by itself lift the exponential dependence on source count. This is why, in the main text, analyses at the scale of many attention heads or many agents necessarily fall back on summary statistics (O-information and pairwise aggregations) rather than the full decomposition; the algorithm here is what those summaries approximate, and what one should run directly whenever the effective number of sources is small (e.g., after grouping heads into a handful of functional modules).

For empirical data the algorithm takes an \emph{estimated} $p_{\bm X,Y}$ as input, so its reliability inherits that of the density estimate. Disclosure atoms, like all PID quantities, are data-hungry: the required sample size grows with the alphabet size $|\supp p_{\bm X}|$, and naive plug-in estimates of the entropies $h_j$ are positively biased, so bias correction (e.g., Nemenman--Shafee--Bialek or Archer--Park--Pillow estimators) is advisable and should be reported alongside sample counts. These are precisely the reproducibility details that any empirical application---including ours---must document, and they live at this layer rather than in the definition of the measure. A reference implementation of the full pipeline, including the polytope construction and the backbone terms, is available in the
\texttt{syndisc} package.\footnote{\url{https://github.com/pmediano/syndisc}}

We provide additional actionable recommendations on deciding analysis approaches based on the size of the synergistic system (Table \ref{tab:practical-guidelines}) and design levers (Table \ref{tab:engineering_view}).

\newcommand{\stage}[2]{\multirow{#1}{*}{\rotatebox[origin=c]{90}{\textbf{#2}}}}

\begin{table}[htpb]
\centering
\small
\begin{tabular}{
  c
  >{\raggedright}p{0.18\linewidth}
  >{\raggedright}p{0.33\linewidth}
  >{\raggedright\arraybackslash}p{0.34\linewidth}
}
\toprule
 & \textbf{Question / stage} & \textbf{Recommended approach} & \textbf{Key caveat} \\
\midrule
\stage{7}{Decide} &
What are the parts, target, and timescale? &
Fix parts, target, timescale $\ell$, and redundancy measure on theoretical grounds &
Not given by the theory; claims are bounded by these choices. Substantiate claims by demonstrating robustness across granularities (e.g., heads and experts), orders (e.g., pairs/triplets), and targets (e.g., group error and principal component) \\
\cmidrule(lr){1-4}

\stage{25}{Estimate} &
Small system ($n \leq \sim 5$), full structure needed &
Exact PID, named redundancy measure (BROJA bivariate; $I_{ccs}$/$I_{min}$ discrete) &
Report the measure; check $\geq 2$ for robustness \\
\addlinespace
& Large system, net balance suffices &
Localized/gradient O-information &
Conflates Red/Syn; use per-node, not one global number; report strength measure (S-information or total correlation) to distinguish more synergy from less redundancy; size-matched random baseline to address bias.\\
\addlinespace
& ``Is there any emergence?'' screen &
Emergence-capacity (pairwise $\mathrm{Syn} > 0$); $\Psi_\ell(V) > 0$ (Eq.~4) &
Negative $\Psi$ inconclusive (redundancy subtracted up to $n$ times); consider redundancy-corrected $\Psi(k,q)$ \citep{sas2025improved} \\
\addlinespace
& LLM circuits---which units &
Pairwise/subset synergy $\rightarrow$ per-node ranks (as in the core work) &
Blind to high-order; needs null; add ablation for causal claims \\
\addlinespace
& Dynamical / multi-agent over time &
TDMI-based causal-emergence / $\Phi$ID &
Requires block-shuffled nulls; timescale $\ell$ is a choice; state $\ell$ and report at least one alternative $\ell$. \\
\addlinespace
& Continuous non-Gaussian activations &
Copula normalization or non-parametric estimators &
Gaussian PID may misestimate; report sample-size sensitivity \\
\cmidrule(lr){1-4}

\stage{13}{Report} &
Reporting $\mathrm{Syn}$ estimates &
Establish that findings are generated by interaction, not simple diffusion, shared experimental cues, autocorrelation, estimator bias, finite-samples bias, bias from alphabet size, etc.; report bias-corrected estimate against a suitable null (block-shuffled for time series; row-shuffled or functional otherwise) &
Syn is a residual of large estimated quantities, so error compounds; plug-in MI is positively biased and inflates synergy \\
\addlinespace
& Comparing across conditions, models, or layers & 
Normalize (by $H(\mathrm{target})$ or $I(X;Y)$) or use synergy–redundancy ranks; matched nulls per condition &
Differences in source entropy masquerade as differences in synergy; result depends on null family \\
\cmidrule(lr){1-4}

\stage{10}{Interpret} &
Making causal claims &
Attribute synergy to a design lever; randomize the lever (ablation, persona, ToM prompt, routing $k$) &
Observational synergy $\ne$ interventional synergy; correlational PID alone does not license a causal reading \\
\addlinespace

& Making mechanism claims &
Treat PID atoms as a description of behavior; support mechanism claims by ablation or by fitting models of increasing interaction order and comparing their fit &
High-order behavior $\ne$ high-order mechanism~\cite{rosas2022disentangling} \\

% & Making claims about performance & 
% Locate performance in the system (Syn > 0 locates it in joint production; Syn \approx 0 in individual pieces), localizing performance in Red/Syn balance informs design decisions and predicts scaling and robustness

% \addlinespace
% TODO: something about tying Syn > 0 to performance
% & Tying $\mathrm{Syn} > 0$ to performance & 
% Pair PID with an outcome model to answer what synergy changes, functionally &
% Synergy can be chaotic or useless (see control condition  was strongly synergistic and uncoordinated. Syn is necessary for emergence claims but says nothing about function.\\

% Inferring emergence from performance gain, or benefit from Syn > 0 — Treat as separate axes and report both — Caveat: gain without Syn is substitution or redundant averaging; Syn without gain is differentiation without alignment (§3.2 control). Neither axis alone identifies productive emergence; the §3.2 Syn × Red interaction is what does.

\bottomrule
\end{tabular}
\vspace{4pt}
\caption{Practical guidelines for applying information decomposition across the research workflow.}
\label{tab:practical-guidelines}
\end{table}

\clearpage 
\newtcolorbox[auto counter, number freestyle={S\noexpand\arabic{\tcbcounter}}]{reportingbox}[1][]{%
  enhanced, breakable,
  width=\linewidth,
  colback=white,
  colframe=darkblue!60,
  colbacktitle=darkblue!10,
  coltitle=black,
  fonttitle=\bfseries\small\sffamily,
  arc=1.5mm,
  boxrule=1pt,
  top=1mm, bottom=1mm, left=1mm, right=1mm,
  boxsep=4pt,
  before skip=10pt, after skip=10pt,
  toptitle=1.2pt,
  bottomtitle=-1.2pt,
  title={Box~\thetcbcounter: Reporting guidelines for synergy claims},
  #1}

\begin{reportingbox}[label={box:reporting}]
\footnotesize
\setlength{\parskip}{2pt}
\noindent Claims about synergy or emergence can be compared and disputed only if the choices that bound it are visible. Table~\ref{tab:practical-guidelines} recommends which analysis to run; this box outlines how to report them.

\smallskip
\noindent\begin{minipage}[t]{0.48\linewidth}
\textbf{A.\ The claim}
\begin{enumerate}[leftmargin=1.3em, itemsep=0pt, topsep=1pt]
  \item \textbf{Quantity and sign:} exact quantity, expression, sign convention.
  \item \textbf{Order:} $n$, order $k$ detected, how larger systems were reduced and aggregated (rule and range across subsets).
  \item \textbf{Reading:} observational vs.\ interventional; behavior vs.\ mechanism; highest rung reached (detect $\rightarrow$ localize $\rightarrow$ identify causally $\rightarrow$ control $\rightarrow$ predict).
  \item \textbf{Scope:} systems, tasks, and scales covered; generality asserted and the cross-context evidence; what the claim does \emph{not} show.
%   \item \textbf{Language:} verbs match the rung and scope; ``proves,'' ``universal,'' and ``fundamental'' need corresponding evidence.
  
\end{enumerate}
\vspace{4pt}
\textbf{B.\ The modeling choices}
\begin{enumerate}[leftmargin=1.3em, itemsep=0pt, topsep=1pt, resume]
  \item \textbf{Parts:} unit, reason for the granularity, one alternative.
  \item \textbf{Target:} $Y$ or macro $V$; fixed, derived, or learned; one alternative.
  \item \textbf{Timescale:} static or dynamical; numeric $\ell$; sensitivity across $\ell$.
  \item \textbf{Redundancy measure:} which, why, non-negativity, a second measure from a different family, able to disagree with the first.
  \item \textbf{Estimator and sample:} data model, residualization, discretisation, $N$ vs.~alphabet and effective $N$ (autocorrelation, mixing), bias correction.
\end{enumerate}
\end{minipage}\hfill
\begin{minipage}[t]{0.48\linewidth}
\textbf{C.\ The comparison}
\begin{enumerate}[leftmargin=1.3em, itemsep=0pt, topsep=1pt, start=10]
  \item \textbf{Null model:} what it preserves and destroys; surrogates; correction; nulls are one-sided.
  \item \textbf{Strength beside balance:} report $\Omega$ or $\Psi_\ell(V)$ next to total information.
  \item \textbf{Whole-minus-sum caveat:} negative $\Psi_\ell(V)$ is inconclusive; a positive $\Psi_\ell(V)$ with a vanishing sum term reflects the macro, not synergy; redundancy is subtracted up to $n$ times.
\end{enumerate}
\vspace{4pt}
\textbf{D.\ The pipeline}
\begin{enumerate}[leftmargin=1.3em, itemsep=0pt, topsep=1pt, resume]
  \item \textbf{Ground truth:} pipeline recovers known atoms on COPY, XOR, AND, and UNQ; the last separates MMI from BROJA.
  \item \textbf{Provenance:} software versions, code, seeds, replicates, sample counts.
\end{enumerate}
\end{minipage}

\smallskip
\noindent\textit{How to use this box.} These are questions a claim should be able to answer, not conditions to be satisfied. Every item has legitimate exceptions; state the reason rather than tick the box. An item that cannot be answered should be reported as unanswered rather than omitted. Reasoned deviation is good practice; silent deviation is what this box is meant to end.
\end{reportingbox}

\clearpage 
\begin{table}[htpb]
\centering
\begin{tabular}{
  >{\raggedright}p{0.15\linewidth}
  >{\raggedright}p{0.25\linewidth}
  >{\raggedright\arraybackslash}p{0.51\linewidth}
}
\toprule
\textbf{PID Principle} & \textbf{Observable Signature} & \textbf{Design Levers} \\
\toprule
Synergy requires differentiation & 
    Identical experts / agents produce only redundancy & 
    Sparsity and routing in MoE systems (larger $k$ in top-$k$ gating $\uparrow$~synergy \citep{urbina2026brain,fulker2024spontaneous});
        data augmentation ($\uparrow$~redundancy \citep{tian2020makes});
        personas/roles ($\uparrow$~synergy \citep{riedl2026emergent});
        division of labor ($\uparrow$ uniqueness) vs.~co-creation ($\uparrow$~synergy)\\
\midrule

Synergy is fragile, redundancy is robust & 
    Sharp failure modes vs.~graceful degradation; 
        flat-gradient stalling on synergistic tasks;
        robust OOD when structure is stable vs.~brittle if composition shifts;
        path-dependent collapse & 
    Dropout ($\uparrow$~redundancy \citep{achille2018information,proca2024synergistic}); 
        BatchNorm / LayerNorm ($\uparrow$~synergy potential \citep{daneshmand2021batch}); 
        early stopping ($\uparrow$~redundancy) vs.~extended training, warmup schedules ($\uparrow$~synergy);
        individual rewards ($\uparrow$~independence \& uniqueness) vs.~shared rewards ($\uparrow$~synergy);
        shaped rewards for coordination ($\uparrow \uparrow$~synergy) \\
\midrule

    Synergistic information is non-local & 
    Not locally attributable; modular interpretability fails & 
    Attention mechanism ($\uparrow$~synergy \citep{urbina2026brain}); 
        data curation in multi-modal training ($\uparrow$~synergy \citep{liang2023quantifying});
        centralized ($\uparrow$~synergy) vs.~decentralized training execution ($\uparrow$~redundancy);
        adaptivity \& personalization ($\uparrow$~synergy \citep{christiano2017deep,lau2024personalized,kelley2025personalized})\\
\midrule

    Synergy provides superlinear capacity & 
    Superlinear performance gains but data-hungry estimation & 
    Network width ($\downarrow$~synergy) vs.~depth ($\uparrow$~synergy); 
        ToM in multi-agent systems \& human-AI ($\uparrow$~synergy; $\uparrow$ differentiation \citep{riedl2021quantifying,riedl2026quantifying,riedl2026emergent});
        full observability ($\downarrow$~synergy) vs.~partial observability ($\uparrow$~synergy)\\
\bottomrule
\addlinespace[4pt]
\end{tabular}

\caption{Overview of control parameters and their effects across layers.}
\label{tab:engineering_view}
\end{table}

\subsection{Summary of evidence across worked examples}
\begin{table}[htpb]
\centering
\begin{tabular}{
  >{\raggedright}p{0.12\linewidth}
  >{\raggedright}p{0.25\linewidth}
  >{\raggedright}p{0.25\linewidth}
  >{\raggedright\arraybackslash}p{0.25\linewidth}
}
\toprule
\textbf{Rung} & \textbf{Synergistic core} \citep{urbina2026brain} & \textbf{Multi-agent} \citep{riedl2026emergent} & \textbf{Human-AI} \citep{riedl2026quantifying} \\
\toprule
Detect & 
    Pairwise $\Phi$ID synergy ranks & 
    $\Psi > 0$; emergence capacity $> 0$ & 
    Bounded residual; formal PID $\mathrm{Syn} > 0$ \\
\midrule

Localize & 
    Middle layers; heads and experts & 
    Agent pairs, not triplets; identity-linked roles & 
    Hard and AI-unsolvable items; $\kappa$ trait \\
\midrule

Identify causally & 
    Ablation of synergistic vs.~redundant heads & 
    Randomized $\cplain$/$\cpersona$/$\ctom$ conditions & 
    Randomized assignment; users $\times$ items crossed; phases rule out learning \\
\midrule

Control & 
    RL fine-tuning of synergistic regions & 
    ToM prompt as control parameter & 
    Open: ToM is measured, not manipulated \\
\midrule

Predict & 
    Conjectures~1--4 & 
    Conjecture~5 & 
    Conjecture~5 \\
\bottomrule
\addlinespace[4pt]
\end{tabular}

\caption{Evidence ladder across the three empirical settings.}
\label{tab:evidence_ladder}
\end{table}

\clearpage 
\subsection{Summary of PID-related software}
\begin{table}[ht]
\centering
\small
\setlength{\tabcolsep}{5pt}
\begin{tabular}{lllccll}
\toprule
\textbf{Name} & \textbf{License} & \textbf{Language} & \textbf{Cont.} & \textbf{Disc.} & \textbf{Framework} & \textbf{Ref.} \\
\midrule
\multicolumn{7}{l}{\textit{Partial Information Decomposition}} \\[2pt]
\texttt{IDTxl}
  & GPLv3 & Python & \checkmark & \checkmark & PID, PED & \cite{wollstadt2018idtxl} \\
\texttt{dit}
  & BSD-3 & Python & \texttimes & \checkmark & PID & \cite{james2018dit} \\
\texttt{BROJA\_2PID}
  & Apache~2 & Python & \texttimes & \checkmark & PID (bivariate) & \cite{makkeh2018broja} \\
\texttt{SxPID}
  & GPLv3 & Python & \texttimes & \checkmark & PID & \cite{makkeh2021introducing} \\
\texttt{infotheory}
  & MIT & Python/C++ & \checkmark & \checkmark & PID & \cite{candadai2019infotheory} \\
\parbox[t]{2.8cm}{\texttt{integrated-info-\newline decomp}}
& BSD-3 & Python & \checkmark & \checkmark & $\Phi$ID & \cite{luppi2024information} \\
\texttt{Informeasure}
  & Artistic-2.0 & R & \texttimes & \checkmark & PID (bivariate) & \cite{pan2024informeasure} \\
\midrule
\multicolumn{7}{l}{\textit{Higher-order interaction summaries}} \\[2pt]
\texttt{HOI}
  & BSD-3 & Python (JAX) & \checkmark & \checkmark & O-info, TC, DTC, S-info & \cite{neri2024hoi} \\
\texttt{THOI}
  & MIT & Python (PyTorch) & \checkmark & \texttimes & O-info, TC, DTC, S-info & \cite{belloli2025thoi} \\
\bottomrule
\end{tabular}

\vspace{4pt}
\begin{flushleft}
\footnotesize
URLs: \texttt{IDTxl}: \url{https://github.com/pwollstadt/IDTxl};
\texttt{dit}: \url{https://github.com/dit/dit};
\texttt{BROJA\_2PID}: \url{https://github.com/Abzinger/BROJA_2PID};
\texttt{SxPID}: \url{https://github.com/Abzinger/SxPID};
\texttt{infotheory}: \url{https://github.com/madvn/infotheory};
\texttt{integrated-info-decomp}: \url{https://github.com/Imperial-MIND-lab/integrated-info-decomp};
\texttt{HOI}: \url{https://github.com/brainets/hoi};
\texttt{THOI}: \url{https://github.com/Laouen/THOI}.
\end{flushleft}

\caption{Non-exhaustive list of available software packages for computing information decomposition and related quantities. Support spans popular environments including Python and R. Runtime and scalability depend on both the entropy estimator (e.g., Gaussian approximation, non-parametric methods) and the specific decomposition. Packages are grouped by scope: full PID implementations, and scalar higher-order interaction summaries (e.g., O-information).}
\label{tab:software}

\end{table}

\clearpage 
\subsection{Testable predictions}
\newcommand{\gap}{\addlinespace[7pt]}

\begin{table}[htpb]
\centering
\begin{tabular}{
  >{\raggedright\arraybackslash}p{.95\linewidth}
}
\toprule
\textbf{Hypothesis} \\
\toprule
\textbf{Grokking as redundancy-to-synergy transition:}\\ 
Track O-information $\Omega$ of the hidden representation across training; confirmed if $\Omega$ flips from positive to negative within the test-accuracy jump (and an $\Omega$-penalty delays it), falsified if $\Omega$ drifts smoothly with train accuracy.\\
\gap

\textbf{Jagged intelligence as a factored-representation boundary:}\\
Score each task by synergistic depth $\mathrm{Syn}(F_1, ..., F_k; Y)$; confirmed if accuracy falls with depth after partialling out human-rated difficulty, falsified if accuracy is unrelated to $\mathrm{Syn}$ or difficulty dominates once depth is controlled.\\
\gap

\textbf{Test-time compute as synergy unrolling:}\\ 
Measure per-pass synergy (synergy-redundancy rank) for a reasoning vs.~base model; confirmed if the reasoning model shows lower per-pass synergy and compute gains shrink with synergistic depth, falsified if its per-pass synergy is equal or higher.\\
\gap

\textbf{Multi-turn degradation and path-dependent synergy collapse:}\\
Regress the single-turn-minus-multi-turn accuracy drop $\Delta$ on synergistic depth; confirmed if $\Delta$ rises with depth and is zero for redundancy-dominated tasks ($\Omega > 0$), falsified if $\Delta$ is uncorrelated with depth.\\
\gap

\textbf{Theory of mind as a cross-level control parameter:}\\
Vary input ToM demand (e.g., via CausalToM from \citep{prakash2026language}) and measure synergy-redundancy rank of the synergistic-core heads; confirmed if high-ToM inputs raise core synergy but not the periphery, falsified if core synergy is insensitive to ToM demand.\\

\bottomrule
\addlinespace[4pt]
\end{tabular}

\caption{Hypotheses in testable form.}
\label{tab:testable}
\end{table}

\end{document}